\documentclass{article}

\usepackage[final]{to_arxiv}

\usepackage[utf8]{inputenc}
\usepackage[T1]{fontenc}
\usepackage{hyperref}
\usepackage{url}
\usepackage{subfigure}
\usepackage{booktabs}
\usepackage{amsfonts}
\usepackage{nicefrac}
\usepackage{microtype}
\usepackage{amsmath}
\usepackage{amsthm}
\usepackage{amssymb}
\usepackage{amscd}
\usepackage{mathtools}
\usepackage{graphicx}
\usepackage{subfigure}
\usepackage{enumitem}
\usepackage[export]{adjustbox}
\mathtoolsset{showonlyrefs=true}
\usepackage[ruled,vlined,linesnumbered]{algorithm2e}

\usepackage{color}
\usepackage{wrapfig}

\title{Leveraging Demonstrations for Deep Reinforcement Learning on Robotics Problems with Sparse Rewards}

\author{
  \textbf{
  Mel Vecerik,
  Todd Hester,
  Jonathan Scholz,
  Fumin Wang} \\
  \textbf{
  Olivier Pietquin,
  Bilal Piot,
  Nicolas Heess} \\
  \textbf{
  Thomas Rothörl,
  Thomas Lampe,
  Martin Riedmiller} \\
  Deepmind\\
  \texttt{vec, toddhester, jscholz, awaw} \\
  \texttt{pietquin, piot, heess} \\
  \texttt{tcr, thomaslampe, riedmiller@google.com} \\
}

\begin{document}
\maketitle

\begin{abstract}
We propose a general and model-free approach for Reinforcement Learning (RL) on real robotics with sparse rewards. We build upon the Deep Deterministic Policy Gradient (DDPG) algorithm to use demonstrations. Both demonstrations and actual interactions are used to fill a replay buffer and the sampling ratio between demonstrations and transitions is automatically tuned via a prioritized replay mechanism. Typically, carefully engineered shaping rewards are required to enable the agents to efficiently explore on high dimensional control problems such as robotics. They are also required for model-based acceleration methods relying on local solvers such as iLQG (e.g. Guided Policy Search and Normalized Advantage Function). The demonstrations replace the need for carefully engineered rewards, and reduce the exploration problem encountered by classical RL approaches in these domains. Demonstrations are collected by a robot kinesthetically force-controlled by a human demonstrator. Results on four simulated insertion tasks show that DDPG from demonstrations out-performs DDPG, and does not require engineered rewards. Finally, we demonstrate the method on a real robotics task consisting of inserting a clip (flexible object) into a rigid object.
\end{abstract}


\section{Introduction}

The latest generation of collaborative robots are designed to eliminate cumbersome path programming by allowing humans to \emph{kinesthetically guide} a robot through a desired motion.
This approach dramatically reduces the time and expertise required to get a robot to solve a novel task, but there is still a fundamental dependence on scripted trajectories.
Consider the task of inserting a wire into a connector: it is difficult to imagine any predefined motion which can handle variability in wire shape and stiffness.
To solve these sorts of tasks, it is desirable to have a richer control \textit{policy} which considers a large amount of feedback including states, forces, and even raw images.
Reinforcement Learning (RL) offers, in principle, a method to learn such policies from exploration, but the amount of actual exploration required has prohibited its use in real applications.
In this paper we address this challenge by combining the demonstration and RL paradigms into a single framework which uses kinesthetic demonstrations to guide a deep-RL algorithm.
Our long-term vision is for it to be possible to provide a few minutes of demonstrations, and have the robot rapidly and safely learn a policy to solve arbitrary manipulation tasks.

The primary alternative to demonstrations for guiding RL agents in continuous control tasks is \textit{reward shaping}.
Shaping is typically achieved using a hand-coded function, such as Cartesian distance to a goal site, which provides a smoothly varying reward signal for every state the agent visits.
While attractive in theory, reward shaping can lead to bizarre behavior or premature convergence to local minima, and in practice requires considerable engineering and experimentation to get right~\cite{ng1999policy}.  By contrast, it is often quite natural to express a task goal as a \textit{sparse} reward function, e.g. +1 if the wire is inserted, and 0 otherwise.
Our central contribution is to show that off-policy replay-memory-based RL (e.g. DDPG) is a natural vehicle for injecting demonstration data into sparse-reward tasks, and that it obviates the need for reward-shaping.
In contrast to on-policy RL algorithms, such as classical policy gradient, DDPG can accept and learn from arbitrary transition data.
Furthermore, the replay memory allows the agent to maintain these transitions for long enough to propagate the sparse rewards throughout the value function.

We present results of simulation experiments on a set of robot insertion problems involving rigid and flexible objects.
We then demonstrate the viability of our approach on a real robot task consisting of inserting a clip (flexible object) into a rigid object.
This task is realized by a Sawyer robotic arm, using demonstrations collected by kinesthetically controlling an arm by the wrist.
Our results suggest that sparse rewards and a few human demonstrations are a practical alternative to shaping for teaching robots to solve challenging continuous control tasks.

\section{Background}

This section provides mathematical background for Markov Decision Processes (MDPs), DDPG, and deep RL techniques such as prioritized replay and $n$-step return.
We adopt the standard Markov Decision Process (MDP) formalism for this work~\cite{SuttonBarto:1998}. An MDP is defined by a tuple $\left<S, A, R, P, \gamma\right>$, which consists of a set of states $S$, a set of actions $A$, a reward function $R(s,a)$, a transition function
$P(s'|s,a)$, and a discount factor $\gamma$.
In each state $s \in S$, the agent takes an action $a \in A$.
Upon taking this action, the agent receives a reward $R(s,a)$ and reaches a new state
$s'$, determined from the probability distribution $P(s'|s,a)$.
A deterministic and stationary policy $\pi$ specifies for each state which action the agent will take.
The goal of the agent is to find the policy $\pi$ mapping states
to actions that maximizes the expected discounted total reward over
the agent's lifetime.
This concept is formalized by the action value function: $Q^\pi(s,a)=\mathbb{E}^\pi\left[\sum_{t=0}^{+\infty}\gamma^tR(s_t,a_t)\right]$, where $\mathbb{E}^\pi$ is the expectation over the distribution of the admissible trajectories $(s_0,a_0,s_1, a_1,\dots)$ obtained by executing the policy $\pi$ starting from $s_0=s$ and $a_0=a$.
Here, we are interested in continuous control problems, and take an actor-critic approach in which both components are represented using neural networks.  These methods consist in maximizing a mean value $J(\theta)=\mathbb{E}_{s\sim\mu}[Q^{\pi(.|\theta)}(s,\pi(s|\theta))]$ with respect to parameters $\theta$ that parameterise the policy and where $\mu$ is an initial state distribution.
To do so, a gradient approach is considered and the parameters $\theta$ are updated as follows: $\theta\leftarrow\theta+\alpha\nabla_\theta J(\theta)$.
Deep Deterministic Policy Gradient (DDPG)~\cite{lillicrap2016continuouss} is an actor-critic algorithm which directly uses the gradient of the Q-function w.r.t. the action to train the policy.
DDPG maintains a parameterized policy network $\pi(.|\theta^\pi)$ (actor function) and a parameterized action-value function network (critic function) $Q(.|\theta^Q)$. It produces new transitions $e=(s,a,r=R(s,a),s'\sim P(.|s,a))$ by acting according to $a=\pi(s|\theta^\pi)+\mathcal{N}$ where $\mathcal{N}$ is a random process allowing action exploration. Those transitions are added to a replay buffer $B$.
To update the action-value network, a one-step off-policy evaluation is used and consists of minimizing the following loss:
\begin{equation}
L_1(\theta^Q)=\mathbb{E}_{(s,a,r,s')\sim D}\left[R_1-Q(s,a|\theta^Q)\right]^2,
\end{equation}
where $D$ is a distribution over transitions $e=(s,a,r=R(s,a),s'\sim P(.|s,a))$ contained in a replay buffer and the one-step return $R_1$ is defined as: $R_1=r+\gamma Q'(s',\pi'(s')|\theta^{\pi'})|\theta^{Q'})$.

Here $Q'(.|\theta^{Q'})$ and $\pi'(.|\theta^{\pi'})$ are the associated target networks of $Q(.|\theta^{Q})$ and $\pi(.|\theta^{\pi})$ which stabilizes the learning (updated every $N'$ steps to the values of their associated networks).
To update the policy network a gradient step is taken with respect to:
\begin{equation}
\nabla_{\theta^\pi}J(\theta^\pi)\approx \mathbb{E}_{(s,a)\sim D}\left[\nabla_aQ(s,a|\theta^Q)_{|a=\pi(s|\theta^Q)}\nabla_{\theta^\pi}\pi(s|\theta^\pi)\right].
\end{equation}
The off-policy nature of the algorithm allows the use of arbitrary data such as human demonstrations.

Our experiments made use of several general techniques from the deep RL literature which significantly improved the overall performance of DDPG on our test domains.  As we discuss in Sec. \ref{sec:results}, these improvements had a particularly large impact when combined with demonstration data.

\section{DDPG from Demonstrations}

Our algorithm modifies DDPG to take advantage of demonstrations. The demonstrations are of the form of RL transitions: $(s,a,s',r)$.
DDPGfD loads the demonstration transitions into the replay buffer before the training begins and keeps all transitions forever.

DDPGfD uses prioritized replay to enable efficient propagation of the reward information, which is essential in problems with sparse rewards.
Prioritized experience replay~\cite{SchaulQAS16} modifies the agent to sample more important transitions from its replay buffer more frequently.
The probability of sampling a particular transition $i$ is proportional to its priority, $P(i) = \frac{p_i^\alpha}{\sum_k p_k^\alpha}$, where $p_i$ is the priority of the transition.
DDPGfD uses $p_i = \delta_i^2 + \lambda_3 \lvert \nabla_aQ(s_i,a_i|\theta^Q) \rvert^2 + \epsilon + \epsilon_D $, where $\delta_i$ is the last TD error calculated for this transition, the second term represents the loss applied to the actor, $\epsilon$ is a small positive constant to ensure all transitions are sampled with some probability, $\epsilon_D$ is a positive constant for demonstration transitions to increase their probability of getting sampled, and $\lambda_3$ is used to weight the contributions.
To account for the change in the distribution, updates to the network are weighted with importance sampling weights, $w_i = (\frac{1}{N} \cdot \frac{1}{P(i)})^\beta$. DDPGfD uses $\alpha = 0.3$ and $\beta = 1$ as we want to learn about the correct distribution from the very beginning.
In addition, the prioritized replay is used to prioritize samples between the demonstration and agent data, controlling the ratio of data between the two in a natural way.

A second modification for the sparse reward case is to use a mix of 1-step and n-step returns when updating the critic function. Incorporating n-step returns helps propagate the Q-values along the trajectories.
The $n$-step return loss consists of using rollouts (forward view) of size $n$ of a policy $\pi$ close to the current policy $\pi(.|\theta^{\pi})$ in order to evaluate the action-value function $Q(.|\theta^Q)$.
The idea is to minimize the difference between the action-value at state $(s=s_0,\pi(s)=a_0)$ and the return of a rollout $(s_i,a_i=\pi(s_i), s_i'\sim P(.|s_i,a_i),r_i)_{i=0}^{n-1}$ of size $n$ starting from $(s,\pi(s))$ and following $\pi$.
The $n$-step return has the following form: $R_n = \sum_{i=0}^{n-1}\gamma^ir_i + \gamma^n Q(s'_{n-1},\pi(s'_{n-1});\theta^{Q'})$.
The loss corresponding to this particular rollout is then: $L_n(\theta^Q)= \frac{1}{2} \left(R_n-Q(s,\pi(s)|\theta^Q)\right)^2$.

A third modification is to do multiple learning updates per environment step.
If a single learning update per environment step is used, each transition will only be sampled as many times as the size of the minibatch.
Choosing a balance between gathering fresher data and doing more learning is in general a complicated trade-off.
If our data is stale, the samples from the replay buffer no longer represent the distribution of states our current policy would experience.
This can lead to wrong Q values in states which were not previously visited and potentially cause our policy and values to diverge.
However in our case we require data efficiency and therefore we need to use each transition several times.
In our experiments, we could increase the number of learning updates to $20$ without affecting the per-update learning efficiency.
In practice, we used the value of $40$ which provided a good balance between learning from previous interaction (data efficiency) and stability.

Finally, L2 regularization on the parameters of the actor and the critic networks are added to stabilize the final learning performance.

The final loss can be written as:
\begin{align}
    &L_{Critic}(\theta^Q)= L_1(\theta^Q) + \lambda_1 L_n(\theta^Q) +  \lambda_2 L^C_{reg}(\theta^Q)
    \\
    & \nabla_{\theta^\pi}L_{Actor}(\theta^\pi) = - \nabla_{\theta^\pi}J(\theta^\pi) + \lambda_2 \nabla_{\theta^\pi}L^A_{reg}(\theta^\pi)
\end{align}

To summarize, we modified the original DDPG algorithm in the following ways:
\begin{itemize}[noitemsep,nolistsep]
  \item Transitions from a human demonstrator are added to the replay buffer.
  \item Prioritized replay is used for sampling transitions across both the demonstration and agent data.
  \item A mix of 1-step $L_1(\theta^Q)$ and n-step return $L_n(\theta^Q)$ losses are used.
  \item Learning multiple times per environment step.
  \item L2 regularization losses on the weights of the critic $L_{reg}^C(\theta^Q)$ and the actor $L_{reg}^A(\theta^\pi)$ are used.
\end{itemize}

\section{Experimental setup}
\label{sec:experimental_setup}

Our approach is designed for problems in which it is easy to specify a goal state, but difficult to specify a smooth distance function for reward shaping that does not lead to sub-optimal behavior.
One example of this is insertion tasks in which the goal state for the plug is at the bottom of a socket, but the only path to reach it, and therefore the focus of exploration, is at the socket opening.
While this may sound like a minor distinction, we found in our initial experiments that DDPG with a simple goal-distance reward would quickly find a path to a local minimum on the \emph{outside} of the socket, and fail to ever explore around the opening.

\begin{figure}[htb]
  \centering
  \def\figwidth{0.4\textwidth}
  \subfigure[Peg Insertion Task.] {
    \includegraphics[width=\figwidth]{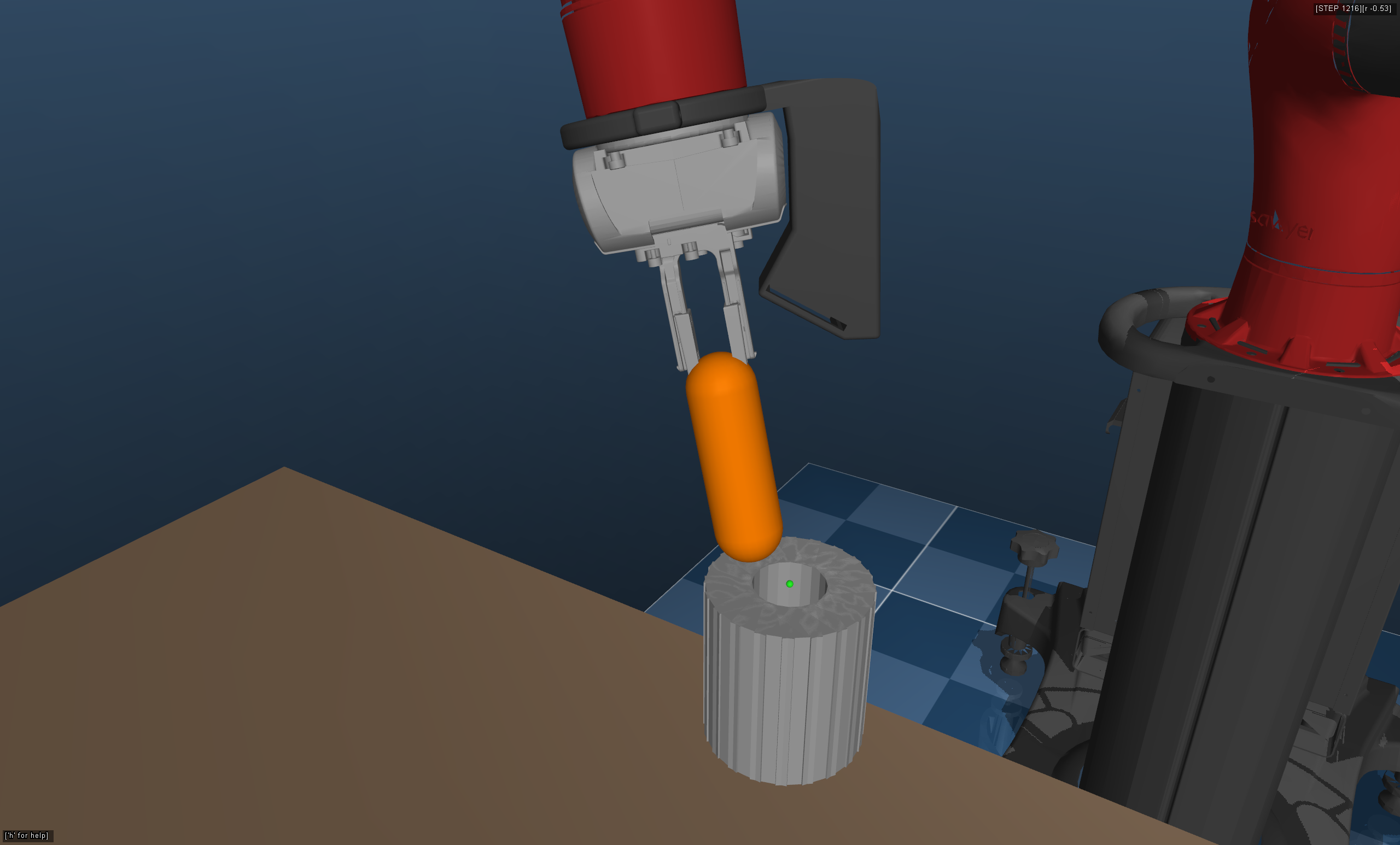}
    \label{fig:peg_task}}
  \subfigure[Hard-drive Task.] {
    \includegraphics[width=\figwidth]{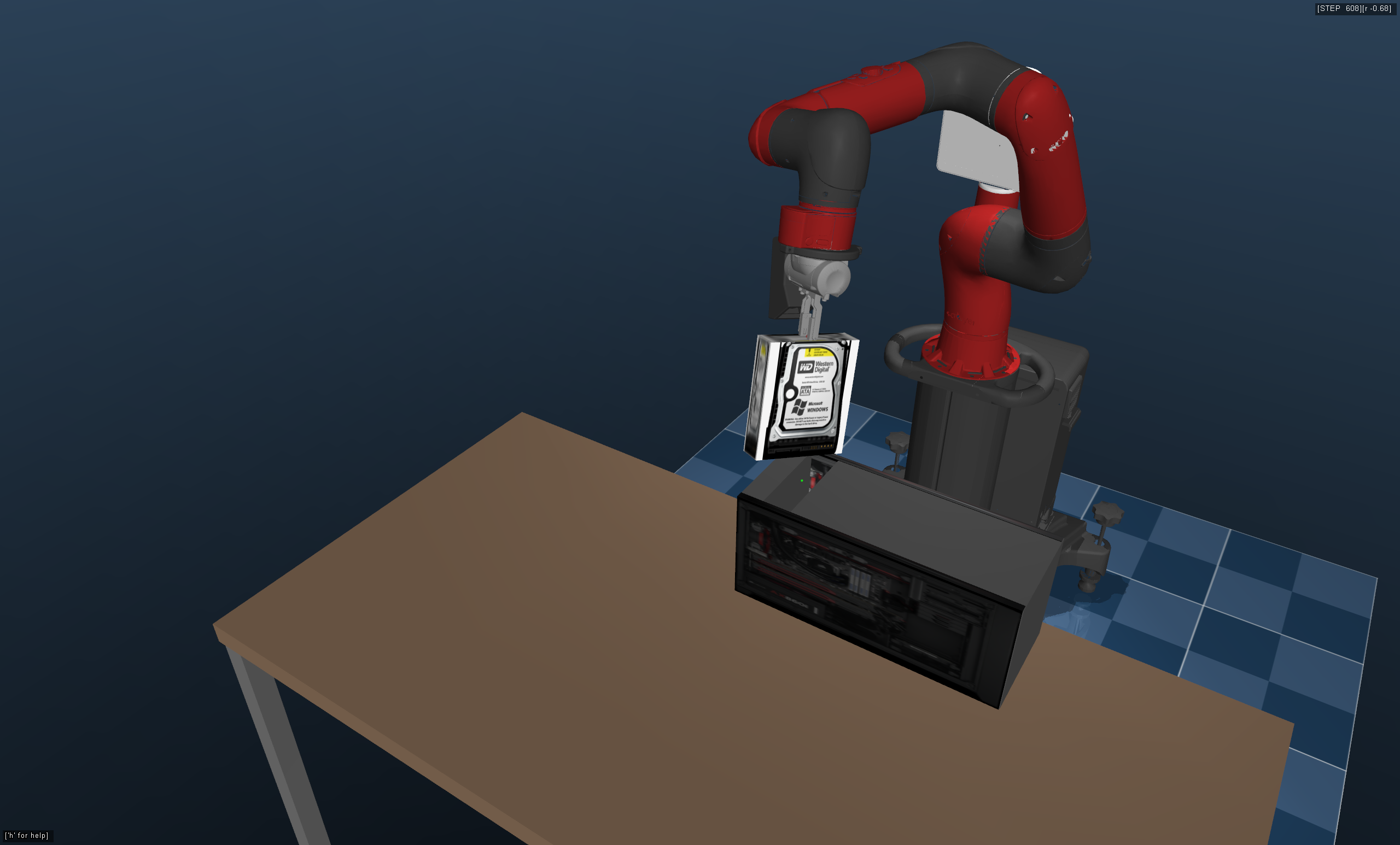}
    \label{fig:harddrive_task}}
  \subfigure[Clip Insertion Task] {
    \includegraphics[width=\figwidth]{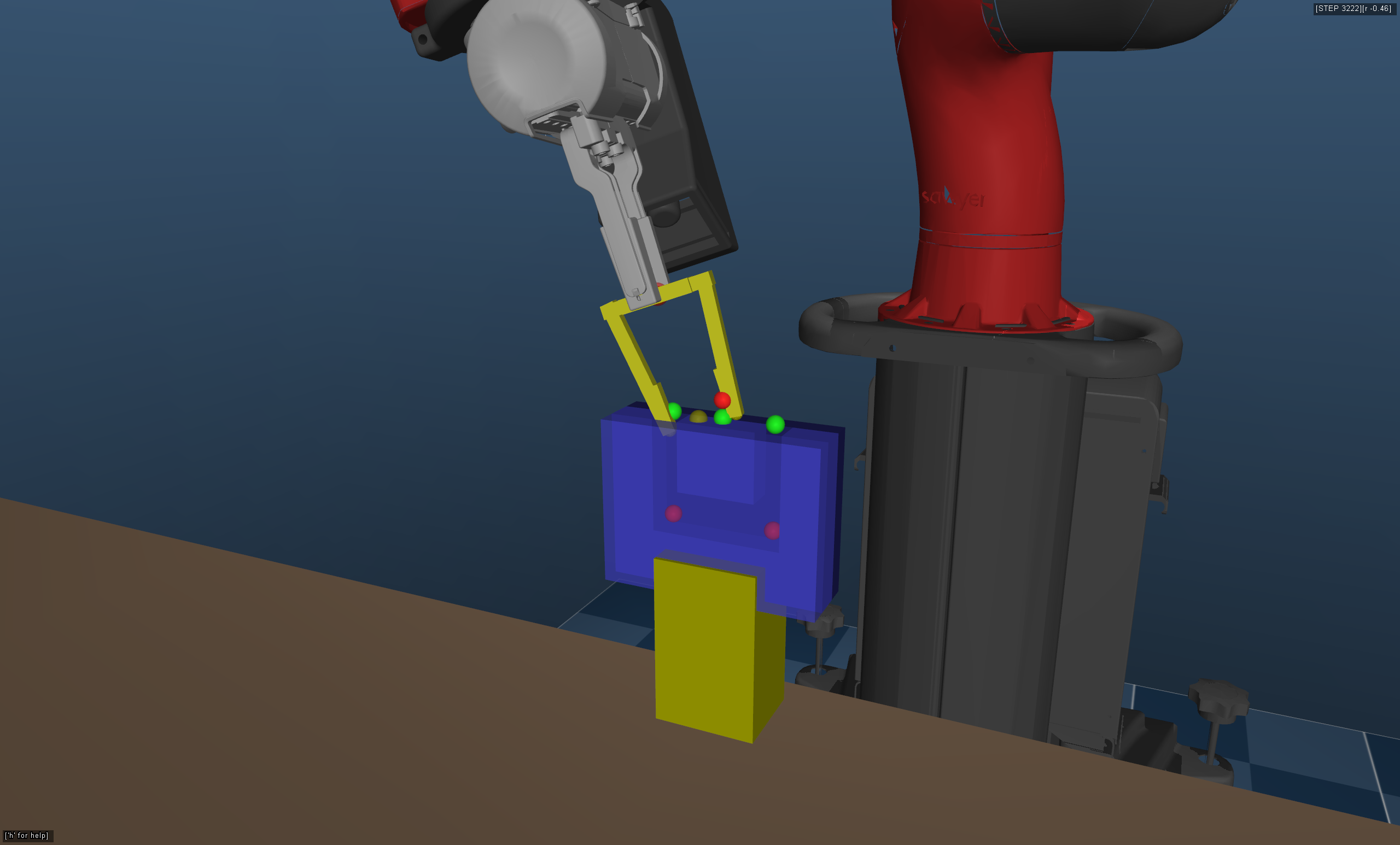}
    \label{fig:clippy_task}}
  \subfigure[Cable Insertion Task.] {
    \includegraphics[width=\figwidth]{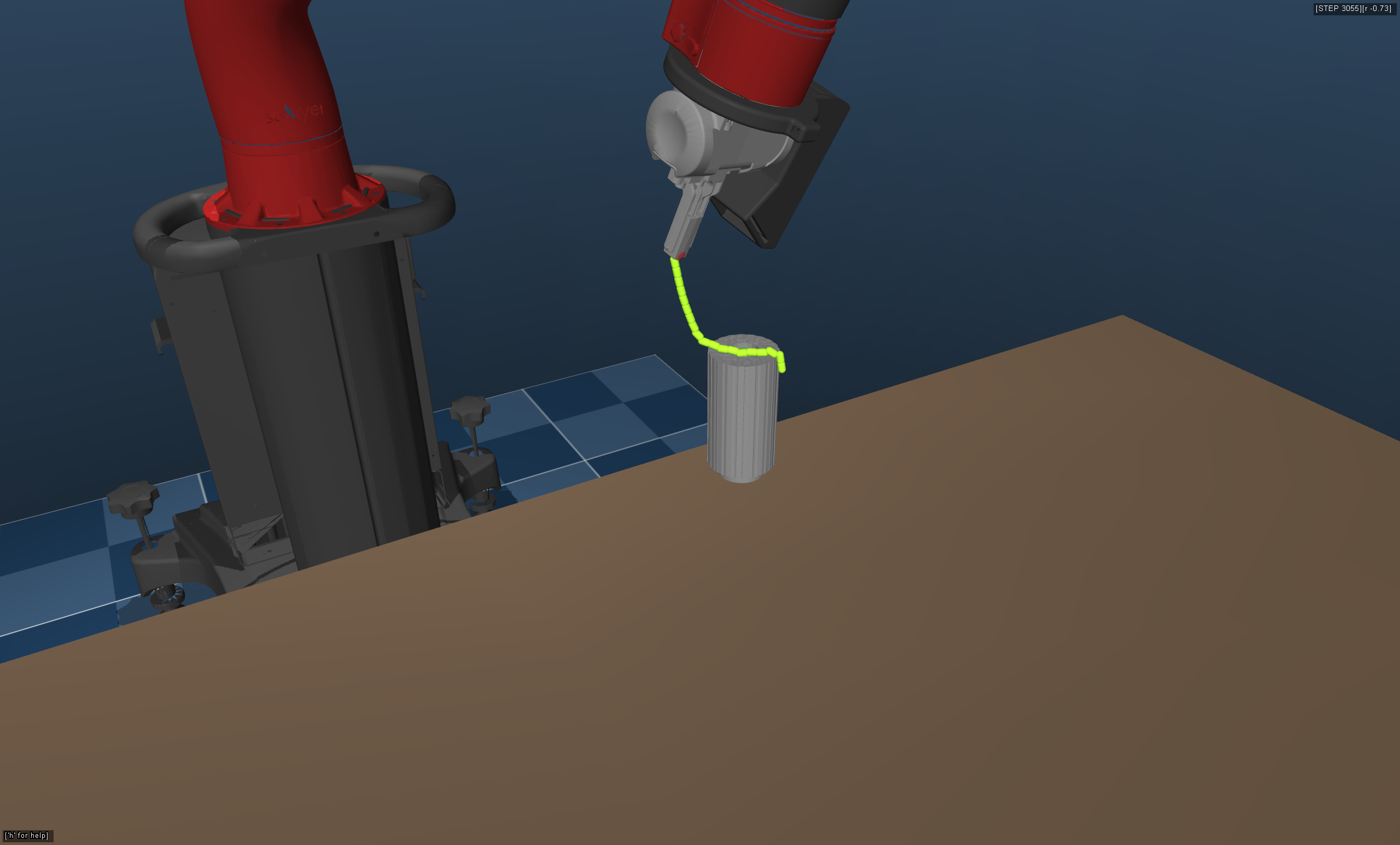}
    \label{fig:cable_task}}
    \vspace{-0.3cm}
  \caption{
  This figure shows the four different insertion tasks.
  }
  \label{fig:tasks}
  \vspace{-0.1cm}
\end{figure}

We therefore sought to design a set of insertion tasks that presented a range of exploration difficulties.  Our tasks are illustrated in Fig. \ref{fig:tasks}.  The first (Fig. \ref{fig:peg_task}) is a classic peg-in-hole task, in which both bodies are rigid, and the plug is free to rotate along the insertion axis.  The second (Fig. \ref{fig:harddrive_task}) models a drive-insertion problem into an ATX-style computer chassis.  Both bodies are again rigid, but in this case the drive orientation is relevant.  The third task (Fig. \ref{fig:clippy_task}) models the problem of inserting a two-pronged deformable plastic clip into a housing.  The clip is modeled as three separate bodies with hinge joints at the base of each prong.  These joints are spring-loaded, and the resting state pinches inwards as is common with physical connectors to maintain pressure on the housing.  The final task (Fig. \ref{fig:cable_task}) is a simplified cable insertion task in which the plug is modeled as a 20-link chain of capsules coupled by ball-joints.  This cable is highly under-actuated, but otherwise shares the same task specification as the peg-in-hole task.

We created two reward functions for our experiments.
The first is a sparse reward function which returned $+10$ if the plug was within a small tolerance of the goal site(s) on the socket:
\vspace{-0.1cm}
\begin{equation}
    r =
    \begin{cases}
     0\ , \quad \sum\limits_{i \in sites} W_g ||g_i - x_i||_2 > \epsilon \\
     10 , \quad \sum\limits_{i \in sites} W_g ||g_i - x_i||_2 < \epsilon
    \end{cases}
    \label{eq:sparse_reward}
\end{equation}
where $x_i$ is the position of the $i^{th}$ tip site on the plug, $g_i$ is the $i^{th}$ goal site on the socket, $W_g$ contains weighting coefficients for the goal site error vector, and $\epsilon$ is a proximity threshold.  If this tolerance was reached, the robot received the reward signal and the episode was immediately terminated.

The second reward function is a shaped reward which composes terms for two movement phases: a reaching phase $c_o$ to align the plug to the socket opening, and an inserting phase $c_g$ to reach the socket goal.  Both terms compute a weighted $\ell_2$-distance between the plug tip(s) and their respective goal site(s). The distance from the goal to the opening site (\emph{i.e.} the maximum value of $c_g$) is added to $c_o$ during the reaching phase, such that the reward monotonically increases throughout an insertion:
\vspace{-0.1cm}
\begin{align}
    c_g &= min\left(\sum\limits_{i \in sites} W_g^T ||g_i-x_i||_2, \sum\limits_{i \in sites} W_o^T ||g_i-o_i||_2\right) \\
    c_o &= I\left(c_g > \sum\limits_{i \in sites} W_o^T ||g_i-o_i||_2\right) \sum\limits_{i \in sites} W_o^T ||o_i-x_i||_2  \\
    r &= \min(1, \max(0, -\alpha \log(\beta (c_o + c_g))) - 1
    \label{eq:shaped_reward}
\end{align}
where $g_i$ is the $i^{th}$ goal site, $o_i$ is the $i^{th}$ opening site, $W_g$ and $W_o$ are weighting coefficients for the goal and opening site errors, respectively, $I$ is the indicator function, and $\alpha$ and $\beta$ are scaling parameters for log-transforming these distances into rewards ranging from $0$ to $1$. Note that tuning the weighting of each dimension in $W_g$ and $W_o$ must be done very carefully for the agent to learn the real desired task. In addition, the shaping of both stages must be balanced out in a delicate manner.

\begin{wrapfigure}{r}{0.45\textwidth}
  \centering
  \def\figwidth{0.45\textwidth}
    \includegraphics[width=\figwidth]{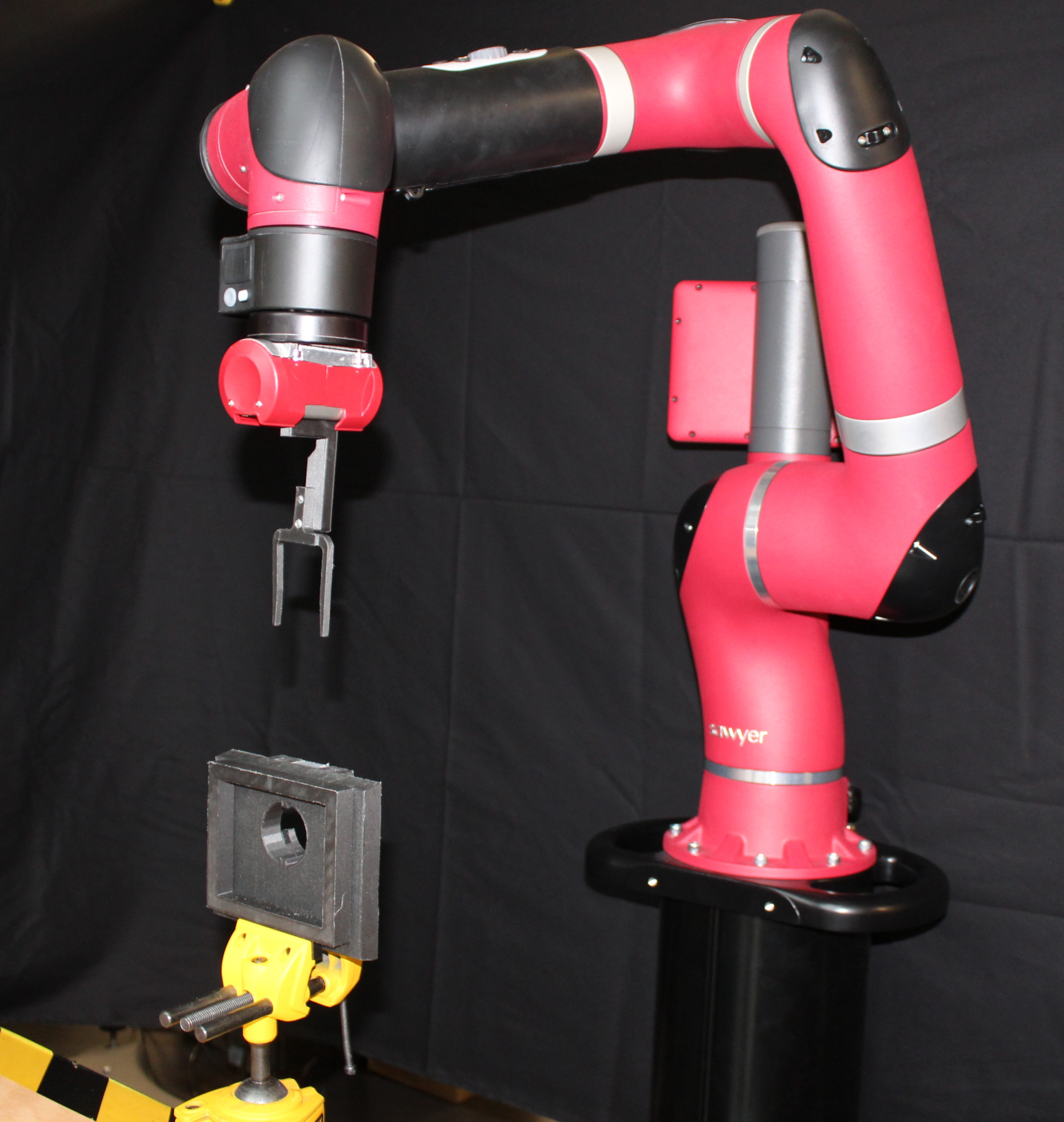}
    \vspace{-0.1cm}
  \caption{Real-robot experiment setup for deformable-clip insertion task.  The clip is made of deformable nylon, and is rigidly attached to the robot gripper.}
  \label{fig:real_sawyer_clippy}
\end{wrapfigure}

All tasks utilized a single vertically mounted robot arm.
The robot was a Sawyer 7-DOF torque-controlled arm from Rethink Robotics, instrumented with a cuff for kinesthetic teaching.
We utilized the Mujoco simulator \cite{todorov2012mujoco} to simulate the Sawyer using publicly available kinematics and mesh files.
In the simulation experiments the actions were joint velocities, the rewards were sparse or shaped as described above, and the observations included joint position and velocity, joint-torque feedback, and the global pose of the socket and plug.
In both the simulation and real world experiments the object being inserted was rigidly attached to the gripper, and the socket was fixed to a table top.

In addition to the four simulation tasks, we also constructed a real world clip insertion problem using a physical Sawyer robot.  In the real robot experiment the clip was rigidly mounted to the robot gripper using a 3D printed attachment.  The socket position was provided to the robot, and rewards were computed by evaluating the distance from the clip prongs (available via the robot's kinematics) to the goal sites in the socket as described above.
In real robot experiments the observations included the robot joint position and velocity, gravity-compensated torque feedback from the joints, and the relative pose of the plug tip sites in the socket opening site frames.

\begin{figure}[t]
  \centering
  \includegraphics[width=\textwidth]{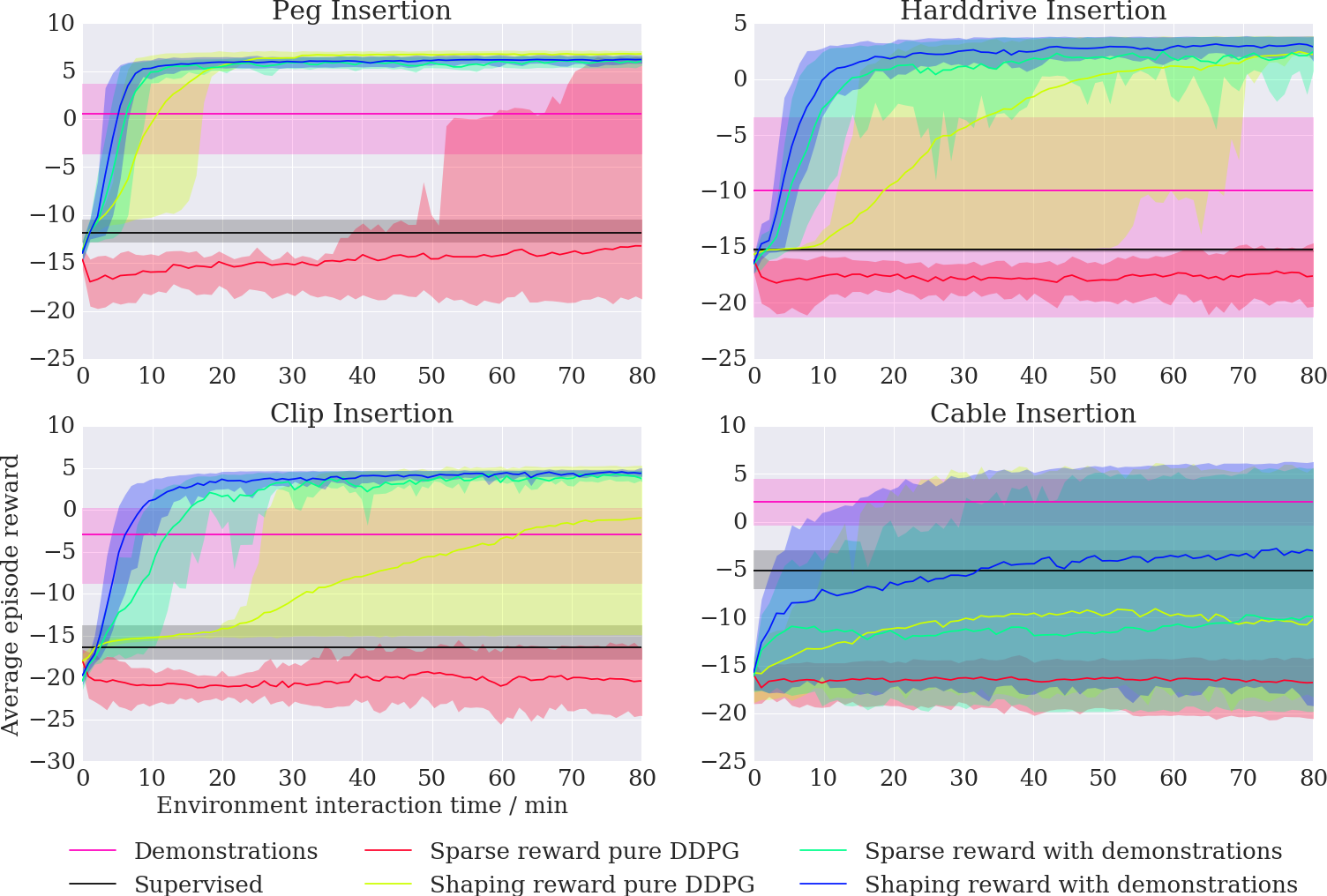}
  \vspace{-0.1cm}
  \caption{
  Learning curves show the means and 10th and 90th percentiles of rewards for the four approaches on each of the four tasks, with statistics computed over 64 trials.
  We measure reward against environment interaction time.
  Each episode was at most 5s long and the agent control rate was about 6Hz.
  The plots also show the mean and percentiles for the rewards received in each set of human demonstration and of supervised imitator which predicts demonstration actions trained with an $\ell_2$ loss.
  The results show that DDPGfD out-performs DDPG, even when DDPG is given hand-tuned shaping rewards and DDPGfD exhibits a more robust training behaviour.
  }
  \label{fig:all_tasks}
  \vspace{-0.2cm}
\end{figure}

\subsection{Demonstration data collection}
To collect the demonstration data in simulated tasks, we used a Sawyer robotic arm.
The arm was kinesthetically force controlled by a human demonstrator.
In simulation an agent was running a hard-coded joint space P-controller to match the joint positions of the simulated Sawyer robot to the joint positions of the real one.
This agent was using the same action space as the DDPGfD agent which allowed the demonstration transitions to be added directly to the agent's replay buffer.

For providing demonstration for the real world tasks we used the same setup, this time controlling a second robotic arm.
Separating the arm we were controlling and the arm which solved the task ensured that the demonstrator did not affect the dynamics of the environment from the agent's perspective.
For each experiment, we collected 100 episodes of human demonstrations which were on average about 25 steps ($\approx5$s) long.
This involved a total of 10-15 minutes of robot interaction time per task.

\section{Results}
\label{sec:results}

In our first experiment we compared our approach to DDPG on sparse and shaped variants of the four simulated robotic tasks presented in Sec. \ref{sec:experimental_setup}. In addition, we show rewards for the demonstrations themselves as well as supervised imitation of the demonstrations.
The DDPG implementation utilized all of the optimizations we incorporated into DDPGfD, including prioritized replay, n-step returns, and $\ell$-2 regularization.
For each task we evaluated the agent with both the shaped and sparse versions of the reward, with results shown in Figure~\ref{fig:all_tasks}.
All traces plot the shaped-reward value achieved, regardless of which reward was given to the agent.  All of these experiments were performed with fixed hyper-parameters, tuned in advance.

We can see that in the case where we have hand-tuned shaping rewards all algorithms can solve the task.
The results show that DDPGfD always out-performs DDPG, even when DDPG is given a well-tuned shaping reward.
In contrast, DDPGfD learns nearly as well with sparse rewards as with shaping rewards. DDPGfD even out-performs DDPG on the hard drive insertion task, where the demonstrations are relatively poor.  In general, DDPGfD not only learns to solve the task, but learns to solve it more efficiently than the demonstrations, usually learning to insert the object in 2-4x fewer steps than the demonstrations.
DDPGfD also learns more reliably, as the percentile plots are much wider for DDPG. Doing purely supervised learning of the demonstration policy performs poorly in every task.

\begin{figure}[t]
  \centering
  \subfigure[Number of demonstration trajectories.] {
    \includegraphics[width=0.5\textwidth,valign=t]{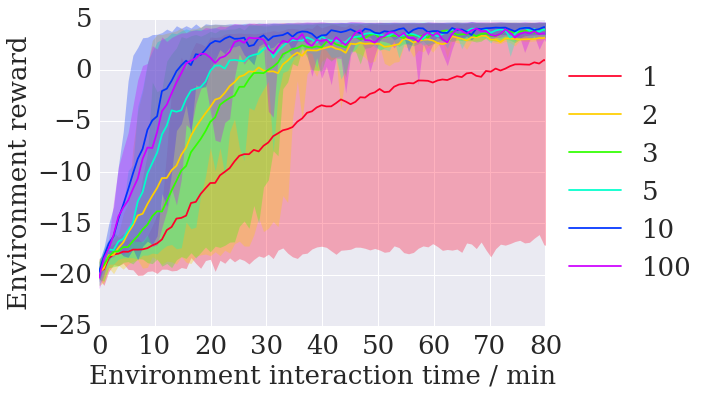}
    \label{fig:dpgfd_num_traces}}
  \subfigure[Real robot experiment.] {
    \includegraphics[width=0.4\textwidth,valign=t]{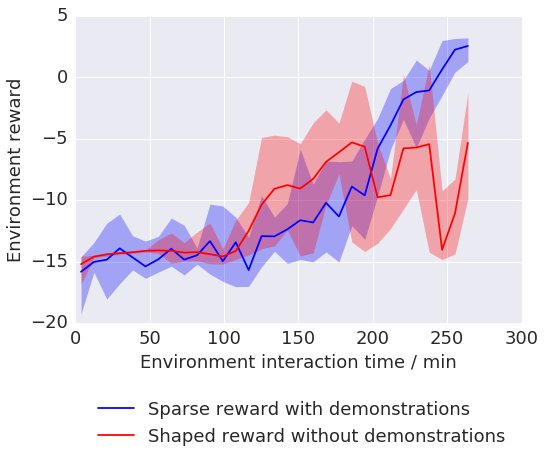}
    \label{fig:real_robot}}
    \vspace{-0.1cm}
  \caption{
  (a) Learning curves for DDPGfD on the clip insertion task with varying amounts of demonstration data. DDPGfD can learn solve the sparse-reward task given only a single trajectory from a human demonstrator. (b) Performance from 2 runs on a real robot. DDPGfD learns faster than DDPG and without the engineered reward function.
  }
  \label{fig:priority}
  \vspace{-0.2cm}
\end{figure}

In our second experiment we examined the effect of varying the quantity of demonstration data on agent performance.
Fig. \ref{fig:dpgfd_num_traces} compares learning curves for DDPGfD agents initialized with 1, 2, 3, 5, 10, and 100 expert trajectories on the sparse-reward clip-insertion task.
DDPGfD is capable of solving this task with only a single demonstration, and we see diminishing returns with 50-100 demonstrations.
This was surprising, since each demonstration contains only one state transition with non-zero reward.

Finally, we show results of DDPGfD learning the clip insertion task on physical Sawyer robot in Figure~\ref{fig:real_robot}. DDPGfD was able to learn a robust insertion policy on the real robot.
DDPGfD with sparse rewards outperforms shaped DDPG, showing that DDPGfD achieves faster learning without the extra engineering.

A video demonstrating the performance can be viewed here:
\url{https://www.youtube.com/watch?v=WGJwLfeVN9w}

\section{Related work}

{\it Imitation learning} is primarily concerned with matching expert demonstrations. Our work combines imitation learning with learning from task rewards, so that the agent is able to improve upon the demonstrations it has seen. Imitation learning can be cast into a supervised learning problem (like classification)~\cite{Pomerleau:1989,ratliff2007}. One popular imitation learning algorithm is DAGGER~\cite{Ross11Dagger} which iteratively produces new policies based on
polling the expert policy outside its original state space. This leads to no-regret over validation data in the online learning sense. DAGGER requires the expert to be available during training to provide additional feedback to the agent.

Imitation can also been achieved through \textit{inverse optimal control} or \textit{inverse RL}. The main principle is to learn a cost or a reward function under which the demonstration data is optimal.
For instance, in \cite{Syed07Game,Syed08Apprenticeship} the inverse RL problem is cast into a two-player zero-sum game where one player chooses policies and the other chooses reward functions.
However, it doesn't scale to continuous state-action spaces and requires knowledge of the dynamics.
To address continuous state spaces and unknown dynamics,~\cite{Klein13} solve inverse RL by combining classification and regression.
Yet it is restricted to discrete action spaces.
Demonstrations have also been used for inverse optimal control in high-dimensional, continuous robotic control problems~\cite{Finn16Guided}.
However, these approaches only do imitation learning and do not allow for learning from task rewards.

Guided Cost Learning (GCL)~\cite{Finn16Guided} and Generative Adversarial Imitation Learning (GAIL)~\cite{Ho16Generative} are the first efficient imitation learning algorithms to learn from high-dimensional inputs without knowledge of the dynamics and hand-crafted features.
They have a very similar algorithmic structure which consists of matching the distribution of the expert trajectories.
To do so, they simultaneously learn the reward and the policy that imitates the expert demonstrations.
At each step, sampled trajectories of the current policy and the expert policy are used to produce a reward function.
Then, this reward is (partially) optimized to produce an updated policy and so on. In GAIL, the reward is obtained from a network trained to discriminate between expert trajectories and (partial) trajectories sampled from a generator (the policy), which is itself trained by TRPO\cite{schulman2015trust}.
In GCL, the reward is obtained by minimization of the Maximum Entropy IRL cost\cite{ziebart2008maximum} and one could use any RL algorithm procedure (DDPG, TRPO etc.) to optimize this reward.

Control in continuous state-action domains typically uses smooth shaped rewards that are designed to be amenable to classical analysis yielding closed-form solutions. Such requirements might be difficult to meet in real world applications. For instance, iterative Linear Quadratic Gaussian (iLQG)~\cite{todorov2005generalized} is a method for nonlinear stochastic systems where the dynamics is known and the reward has to be quadratic (and thus entails hand-crafted task designs). It uses iterative linearization of the dynamics around the current trajectory in order to obtain a noisy linear system (where the noise is a centered Gaussian) and where the reward constraints are quadratic. Then the algorithm uses the Ricatti family of equations to obtain locally linear optimal trajectories that improve on the current trajectory.

Guided Policy Search~\cite{levine2013guided} aims at finding an optimal policy by decomposing the problem into three steps. First, it uses nominal or expert trajectories, obtained by previous interactions with the environment to learn locally linear approximations of its dynamics. Then, it uses optimal control algorithms such as iLQG or DDP to find the locally linear optimal policies corresponding to these dynamics. Finally, via supervised learning, a neural network is trained to fit the trajectories generated by these policies. Here again, there is a quadratic constraint on the reward that must be purposely shaped.

Normalized Advantage Functions (NAF)~\cite{gu2016continuous} with model-based acceleration is a model-free RL algorithm using imagination rollouts coming from a model learned with the previous interactions with the environment or via expert demonstrations. NAF is the natural extension of Q-Learning in the continuous case where the advantage function is parameterized as a quadratic function of non-linear state features. The uni-modal nature of this function allows the maximizing action for the Q-function to be obtained directly as the mean policy.  This formulation makes the greedy step of Q-Learning tractable for continuous action domains. Then, similarly as GPS, locally linear approximations of the dynamics of the environment are learned and iLQG is used to produce model-guided rollouts to accelerate learning.

The most similar work to ours is DQfD~\cite{hester2017learning}, which combines Deep Q Networks (DQN)~\cite{Mnih:2015} with learning from demonstrations in a similar way to DDPGfD. It additionally adds a supervised loss to keep the agent close to the policy from the demonstrations. However DQfD is restricted to domains with discrete action spaces and is not applicable to robotics.

\section{Conclusion}
\label{sec:conclusion}

In this paper we presented DDPGfD, an off-policy RL algorithm which uses demonstration trajectories to quickly bootstrap performance on challenging motor tasks specified by sparse rewards. DDPGfD utilizes a prioritized replay mechanism to prioritize samples across both demonstration and self-generated agent data. In addition, it incorporates n-step returns to better propagate the sparse rewards across the entire trajectory.

Most work on RL in high-dimensional continuous control problems relies on well-tuned shaping rewards both for communicating the goal to the agent as well as easing the exploration problem. While many of these tasks can be defined by a terminal goal state fairly easily, tuning a proper shaping reward that does not lead to degenerate solutions is very difficult. This task only becomes more difficult when you move to multi-stage tasks such as insertion. In this work, we replaced these difficult to tune shaping reward functions with demonstrations of the task from a human demonstrator. This eases the exploration problem without requiring careful tuning of shaping rewards.

In our experiments we sought to determine whether demonstrations were a viable alternative to shaping rewards for training object insertion tasks.
Insertion is an important subclass of object manipulation, with extensive applications in manufacturing.
In addition, it is a challenging set of domains for shaping rewards, as it requires two stages: one for reaching the insertion point, and one for inserting the object.
Our results suggest that Deep-RL is poised to have a large impact on real robot applications by extending the learning-from-demonstration paradigm to include richer, force-sensitive policies.

\clearpage

\bibliography{main}

\newpage
\appendix

\section{Real robot safety}
To be able to run DDPG on the real robot we needed to ensure that the agent will not apply excessive force.
To do this we created an intermediate impedance controller which subjects the agent's commands to safety constraints before relaying them to the robot.
It modifies the target velocity set by the agent according to the externally applied forces.
\begin{equation}
u_{control} = u_{agent} k_a + f_{applied} k_f
\end{equation}
Where $u_{agent}$ is agent's control signal, $f_{applied}$ are externally applied forces such as the clip pushing against the housing, and $k_a$ and $k_f$ are constants to choose the correct sensitivity.
We further limit the velocity control signal $u_{control}$ to limit the maximal speed increase while still allowing the agent to stop quickly.
This increases the control stability of the system.

This allowed us to keep the agent's control frequency, $u_{agent}$, at $5$Hz while still having a physically safe system as $f_{applied}$ and $u_{control}$ were updated at $1$kHz.

\begin{algorithm}[h!]
\DontPrintSemicolon
\SetAlgoLined
\SetKwInOut{Input}{Input}
\SetKwInOut{Output}{Output}
\Input{$Env$ Environment; $\theta^\pi$ initial policy parameters; $\theta^{\pi'}$ initial policy target parameters.}
\Input{$\theta^Q$ initial action-value parameters; $\theta^{Q'}$ initial action-value target parameters; $N'$ target network replacement frequency; $\epsilon$ action noise.}
\Input{$B$ replay buffer initialized with demonstrations; $k$ number of pre-training gradient updates}
\Output{$Q(.|\theta^Q)$ action-value function (critic) and $\pi(.|\theta^\pi)$ the policy (actor).}
\BlankLine

\tcc{Learning via interaction with the environment}
\For{ episode $e\in \{1,\dots,M\}$}{
     \BlankLine
     Initialise state $s_0\sim Env$ \;
     \For{steps $t\in\{1,\dots\, EpisodeLength\}$}{
        Sample noise from Gaussian $n_t = \mathcal{N}(0, \epsilon)$ \;
        Select an action $a_t=\pi(s_{t-1}, \theta^\pi)+n_t$\;
        Get next state and reward $s_t, r_t = T(s_{t-1},a_t), R(s_t)$ \;
        Add single step transition $(s_{t-1}, a_t, r_t, \gamma, s_t)$ to the replay buffer \;
        Add $n$-step transition $(s_{t-n},a_{t-n+1},\sum_{i=0}^{n-1}{r_{t-n+1+i} \gamma^i}, \gamma^n, s_t)$ to the replay buffer \;
    }
     \For{steps $l\in\{1,\dots\, EpisodeLength \times LearningSteps\}$}{
        Sample a minibatch of with prioritization from $D$ and calculate $L_1(\theta^Q)$ and $L_n(\theta^Q)$ as appropriate for a given transition\;
        Update the critic using a gradient step with loss: $L_{Critic}(\theta^Q)$\;
        Update the actor:$\nabla_{\theta^\pi}L_{Actor}(\theta^\pi)$\;
        \If{$step \equiv 0 \pmod{N'}$}{
            Update the target networks: $\theta^{\pi'} \leftarrow \theta^{\pi}$ and $\theta^{Q'} \leftarrow \theta^{Q}$\;
        }
     }
 }
 \caption{DDPG from Demonstrations}
 \end{algorithm}

\end{document}